\documentclass[pdflatex,sn-basic]{sn-jnl}

\usepackage{graphicx}
\usepackage{amsmath,amssymb}
\usepackage{xcolor}
\usepackage{booktabs}
\usepackage{tikz}
\usetikzlibrary{positioning,shapes.symbols}
\usepackage{xurl}
\usepackage{textcomp}

\hypersetup{
  pdftitle={A Critical Analysis of Trustworthy AI Tools, Mark Frameworks, and the Implementation Chasms},
  pdfauthor={Michael Papademas, Xenia Ziouvelou, Kostas Karpouzis, Vangelis Karkaletsis}
}

\newcommand{\refentry}[1]{%
  \par\noindent\hangindent=1.5em\hangafter=1 #1\par\vspace{0.35em}}

\begin{document}

\title[A Critical Analysis of Trustworthy AI Tools]{A Critical Analysis of Trustworthy AI Tools, Mark Frameworks, and the Implementation Chasms}

\author[1,2]{\fnm{Michael} \sur{Papademas}}
\author[1]{\fnm{Xenia} \sur{Ziouvelou}}
\author[2]{\fnm{Kostas} \sur{Karpouzis}}
\author[1]{\fnm{Vangelis} \sur{Karkaletsis}}

\affil[1]{\orgname{Institute of Informatics and Telecommunications },
  \orgdiv{National Centre for Scientific Research ``Demokritos''},
  \orgaddress{\city{Aghia Paraskevi}, \country{Greece}}}

\affil[2]{\orgdiv{Department of Communication, Media and Culture},
  \orgname{Panteion University of Social and Political Sciences},
  \orgaddress{\city{Athens}, \country{Greece}}}

\abstract{As artificial intelligence (AI) systems increasingly impact society, ensuring their ethical and trustworthy deployment has become a global priority. While a myriad of high-level ethical guidelines have emerged, criticism persists that these frameworks remain abstract and lack concrete mechanisms for implementation. This paper conducts a critical analysis of tools and trust mark frameworks intended to operationalize trustworthy AI (TAI), drawing on a comprehensive dataset from the OECD. Through empirical mapping and descriptive comparative analysis, we identify significant asymmetries in ethical focus, lifecycle coverage, stakeholder targeting, and tool typology. Our findings show a strong emphasis on fairness, transparency, and robustness, with comparatively little attention paid to explainability, digital security, and environmental sustainability. Moreover, most tools and certifications concentrate on post-development stages, with limited guidance for early design or data collection phases. Educational initiatives and policy engagement are notably underdeveloped, suggesting that current TAI efforts are dominated by technical and procedural measures within industry contexts. We argue that bridging the persistent chasm between AI principles and practice requires expanding ethical objectives, embedding ethics across the AI lifecycle, and fostering broader multi-stakeholder participation. This study provides both a diagnosis of existing implementation gaps and actionable recommendations for advancing more holistic, inclusive, and enforceable AI governance.}

\keywords{trustworthy AI, ethical AI governance, OECD catalogue, AI lifecycle}

\maketitle

\section{Introduction}\label{sec:introduction}

Ensuring the ethics and trustworthiness of AI systems has become an
ascendant concern in recent years. Many organizations, academic
institutes, industries, and policy sectors have issued AI ethics
principles and guidelines. Over the past decade, more than 80 such
frameworks have been developed since 2016 (Schiff et al. 2020a; Munn
2023), illustrating what has been described as a ``cataclysm'' of
ethical guidelines in artificial intelligence. Across these documents,
there is a notable convergence around a core set of ethical principles.
Systematic literature reviews (SLRs) and meta-analyses have found that
transparency, fairness/justice, non-maleficence/safety,
explicability/accountability, privacy, and
humanity/beneficence/sustainability are the most recurrent values,
appearing in the majority of guidelines (Fjeld et al. 2020; Jobin et al.
2019). This indicates broad agreement that AI systems should be
transparent in operation, treat individuals fairly, be technically and
ethically robust and safe to avoid causing harm, respect privacy, and
have clear accountability mechanisms. These priorities echo the ethical
pillars identified in academic literature and standards efforts
(European Commission 2018; OECD n.d.; European Commission 2020; Laine et
al. 2024).

Paradoxically, despite this high-level consensus, critics have observed
that many AI ethics guidelines remain abstract and ``toothless'', often
articulating lofty principles without providing concrete mechanisms for
implementation (Rességuier and Rodrigues 2020; Munn 2023). The chasm
between principles and practice is widely acknowledged (Papademas et al.
2025). Indeed, even a member of the European expert group criticized the
guidelines as ``lukewarm'' and ``deliberately vague'', asserting that
they relied on rhetoric to gloss over complex challenges such as
explainability (Metzinger 2019). Furthermore, the methods for
operationalizing ethical principles, including the use of technical
tools, organizational processes, and training programs, are often left
unspecified. This lack of clarity leaves AI developers uncertain about
how to effectively translate ethical ideals into practical application
(Rees and Müller 2023; Van den Bergh and Deschoolmeester 2010). This
lack of practical guidance and accountability has raised concerns of
ethicswashing, wherein organizations proclaim principles without
changing behavior (Bietti 2020; Gianni et al. 2022; Schultz et al.
2024).

In response to these problematic aspects, there have been growing
efforts to develop practical tools, standards, and governance mechanisms
to ensure the trustworthiness of AI systems. These include technical
tools, procedural tools, educational resources, as well as certification
and quality mark initiatives that label AI systems as trustworthy if
they meet certain criteria. The OECD has catalogued a wide range of such
trustworthy AI (TAI) tools and trust/quality mark programs. However, to
our knowledge, a systematic analysis of these initiatives, including the
ethical objectives they emphasize, the stages of the AI lifecycle they
address, the audiences they target, and the required skills they use to
support compliance, has been absent from the academic literature.
Addressing this gap, our study draws on the OECD catalogue of AI ethics
tools and trust mark schemes to analyze patterns and disparities within
the current AI trustworthiness ecosystem. By linking these empirical
insights to the existing scholarly discourse, we assess the extent to
which observed trends align with or diverge from established concerns in
AI ethics governance.

This paper begins by outlining the methodology underlying the OECD
catalogue in Section~\ref{sec:methodology}, which serves as the basis for our analysis. We
then present, in Section~\ref{sec:findings}, the key findings on coverage of ethical
objectives, the types of tools used, the lifecycle phases addressed, the
targeted stakeholders, and the expected beneficial outcomes. In Section
4, we contextualize these results within the broader body of existing
research, highlighting their implications for the advancement of more
effective and inclusive AI governance frameworks. We conclude by
summarizing the main insights and offering recommendations for future
research to help bridge the persistent chasm between ethical intentions
and practical outcomes in AI governance.

\section{Methodology}\label{sec:methodology}

This study is based on data collected and categorized by the OECD on
trustworthy AI tools and trust/quality mark frameworks. Specifically, as
of 17 July 2025, a total of 938 tools were identified, among which 24
were categorized as trust/quality marks. The OECD compiled a catalogue
of AI tools\footnote{\url{https://oecd.ai/en/catalogue/tools}} aligned with
the OECD AI Principles as well as a set of AI trustworthiness ``marks''
or certification programs across different organizations. The tools in
the catalogue include technical, educational, and procedural tools. Each
tool is annotated with the specific trustworthy AI objectives it
supports. Likewise, the trust/quality mark initiatives are documented
with details such as the AI system lifecycle stages they address, their
intended target users, required skills for implementation, and stated
objectives and benefits. Our analysis provides quantitative counts or
frequencies for each of these categories, along with a qualitative
interpretation of the quantitative patterns.

We adopt a descriptive and comparative approach to examine this dataset.
First, we aggregate the distribution of tools across the OECD's defined
trustworthy AI objectives. This reveals which ethical principles are
most and least commonly addressed by available tools. We further break
this down by tool type (technical, educational, procedural) to identify
any discrepancies in focus across tool types. Second, we analyze the
trust/quality mark frameworks along multiple dimensions: (a) Lifecycle
stage focus, identifying which stages of the AI system development
lifecycle (e.g., design, data collection, validation,
deployment/monitoring) are most frequently covered by these marks; (b)
Target audience refers to the role or stakeholder groups (e.g.,
developers, data scientists, business leaders, general employees,
policymakers, etc.) the frameworks primarily aim to serve; (c) Required
competencies eliciting what skills or expertise the frameworks assume
implementers must have; (d) Ethical objectives recognizing which
trustworthy AI principles are explicitly the goals of these
certification programs; (e) Expected benefits refer to the positive
outcomes the programs claim to deliver, such as risk reduction or
quality improvement. We interpret frequency counts in each category to identify
dominant trends and gaps. This methodology blends an empirical overview
of current AI ethics implementations (via the OECD's catalogue) with a
scholarly grounding, enabling us to discuss not only the current state
of practice but also why these patterns may be occurring and how they
compare with broader trends noted in the literature. The next section
details the findings, structured by thematic categories derived from our
analytical approach.

This study is subject to certain limitations that should be explicitly
acknowledged. First, the analysis relies on the OECD catalogue as a
pre-existing, externally curated dataset and therefore adopts the
OECD's classifications, labels, and category assignments
without independently recoding the entries. Second, because the study
examines the full set of entries available in the OECD repository at the
time of data collection, it does not employ a sampling strategy. While
this provides a comprehensive descriptive overview of the selected
repository, it also means the results are bounded by the
repository\textquotesingle s scope, inclusion criteria, and potential
selection biases. Third, the present analysis is descriptive and
comparative, focusing on frequencies and distributions across categories
to identify dominant trends, asymmetries, and chasms. These limitations
delimit the study's claims and indicate that the findings should be read
primarily as an empirical mapping of the OECD-curated ecosystem of
trustworthy AI tools and trust/quality mark frameworks.

\section{Synthesis of Key Findings}\label{sec:findings}

\subsection{Emphasis on Core Ethical Principles and Underrepresented Areas}

The distribution of tools across ethical principles is highly
non-uniform. We found a strong concentration of tools addressing a few
key principles, while several other principles receive relatively little
tool support. In particular, transparency is the most widely supported
objective in the OECD's catalogue of trustworthy AI tools. Numerous
tools have been developed to enhance the transparency of AI systems, for
instance, by documenting model decision processes or data provenance.
Following transparency, the next most common goals are fairness and
robustness. This indicates that the AI developer community has invested
substantial effort in developing technical solutions for bias mitigation
(fairness) and model safety and stability (robustness), as well as
procedural frameworks to govern these aspects. These priorities align
with global ethical AI guidelines and frameworks, which frequently
highlight fairness and transparency as fundamental principles (Papademas
et al. 2025). Robustness also appears in most frameworks, confirming
that fairness, transparency, and technical robustness together form a
core focus in both guidance and practical tool development for
trustworthy AI.

\begin{figure}[tbp]
  \centering
  \includegraphics[width=\linewidth]{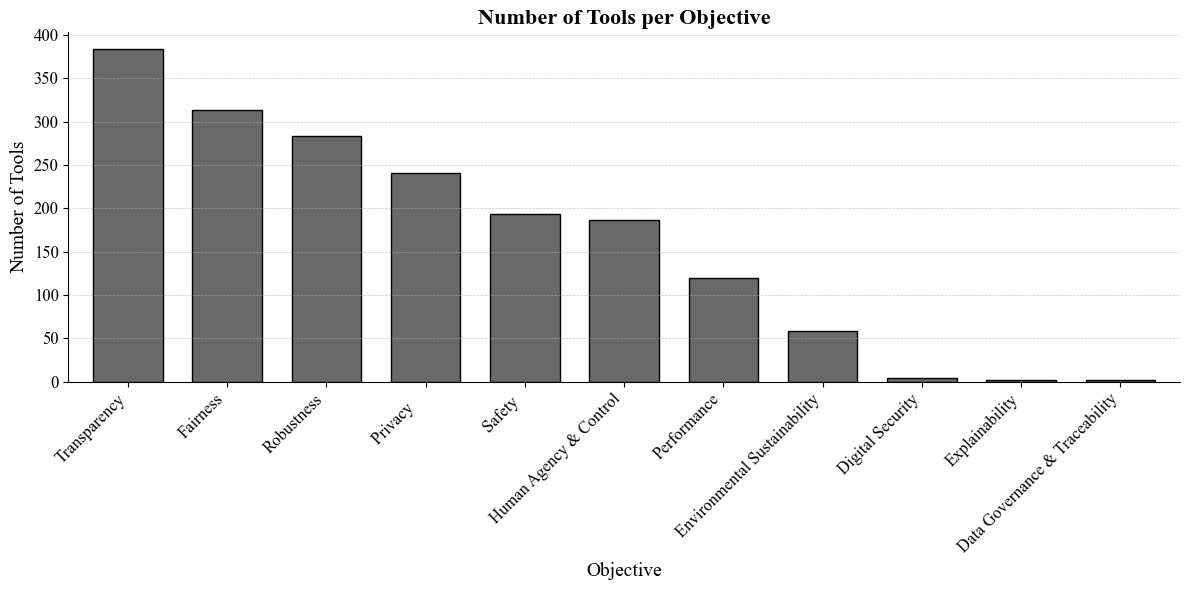}
  \caption{Number of tools in the OECD catalogue per trustworthy-AI objective.}
  \label{fig:tools-by-objective}
\end{figure}

Figure~\ref{fig:tools-by-objective} reveals that several important principles are underserved by
existing tools. Notably, explainability, which refers to the capacity to
interpret and comprehend AI decisions, is addressed far less frequently.
Compared to tools developed for transparency, relatively few explicitly
focus on explainability. This might reflect the greater technical
challenge of providing human-interpretable explanations for complex
models, as well as some ambiguity in how ``explainability'' is defined
distinct from transparency (Floridi and Cowls 2019). Another striking
gap is in digital security: tools focusing on security (e.g., defending
against adversarial attacks or data breaches in AI systems) are
relatively scarce in the OECD catalogue for trustworthy AI. This
suggests that while security is recognized as important, it may not have
seen the same level of tooling. Finally, environmental sustainability
considerations are underrepresented. Very few tools explicitly aim to
address the environmental impact of AI, such as energy efficiency or
carbon footprint. This aligns with prior reviews of AI ethics
guidelines, which report a persistent lack of mention of sustainability
as an ethical principle in many such documents (Corrêa et al. 2023). In
our data, sustainability appears to be an afterthought, indicating a
clear chasm in the current tool ecosystem.

Our findings provide empirical confirmation that this disequilibrium
extends into implementation, with tools available for what is easiest or
most demanded, such as bias audits and transparency documentation, while
areas that are newer, more difficult, or less mandated, like
environmental impact or deep model explainability, lag and remain
underdeveloped. The consequences are that practitioners have ample
resources to tackle certain issues (fairness, transparency), but fewer
tools are available to address other equally important dimensions (e.g.,
how to ensure an AI system is explainable or energy-efficient). This
discrepancy highlights a need identified in the literature to broaden
the scope of ethical AI efforts (Rees and Müller 2023). Researchers have
called for greater incorporation of underrepresented principles such as
sustainability and human dignity into the AI ethics agenda (Rohde et al.
2023; Klein and D'Ignazio 2024; Celsi and Zomaya 2025), and our analysis
of available tools supports that call.

\subsection{Differences Across Categories of Tool Types}

\begin{figure}[tbp]
  \centering
  \includegraphics[width=\linewidth]{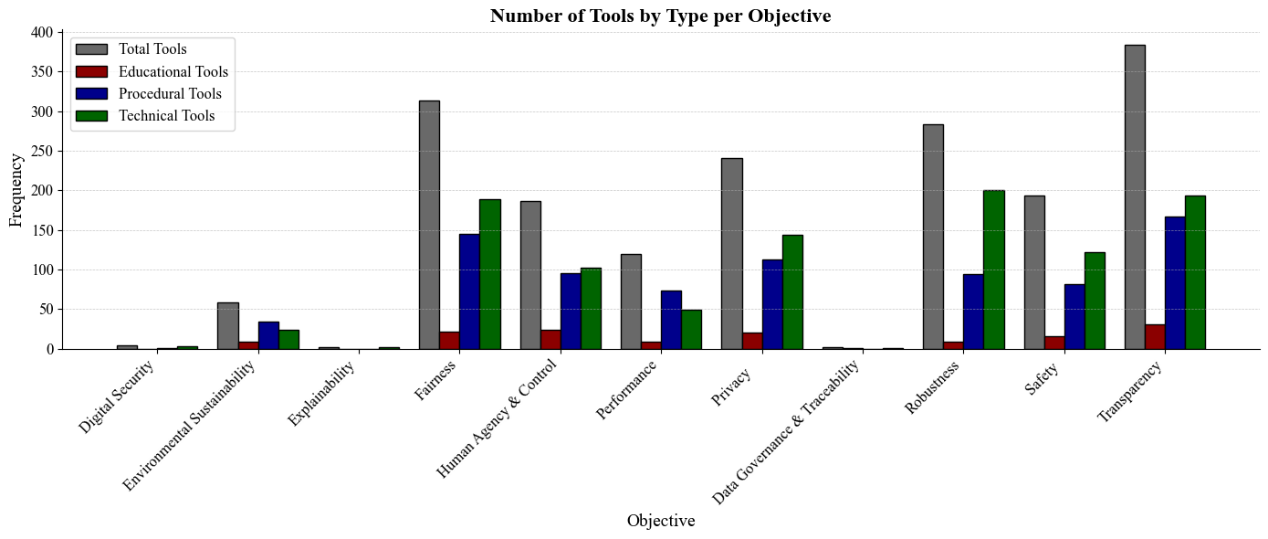}
  \caption{Number of tools by type in the OECD catalogue per trustworthy-AI objective.}
  \label{fig:tools-by-type}
\end{figure}

The OECD data allow a more analytical look at how different categories
of tools address ethical objectives. In Figure~\ref{fig:tools-by-type}, we observe notable
inconsistencies between technical, educational, and procedural tools in
terms of which principles they emphasize. Technical tools that address
aspects such as fairness, transparency, explainability, performance,
robustness, safety, and cybersecurity typically include software
libraries, algorithms, or platforms used directly in the development or
evaluation of AI systems. Our findings show that these tools are heavily
concentrated around a few objectives, primarily transparency,
robustness, and fairness, which together account for the majority of
technical tools in the catalogue. This suggests that innovation and
engineering efforts have focused on more quantifiable areas like
measuring bias or testing model stability and performance. In contrast,
significantly fewer tools target objectives such as explainability or
security, likely because explaining AI decisions or securing systems
against attacks often requires custom, context-specific solutions that
are more difficult to generalize into widely usable libraries. The
relative scarcity of tools in these domains may indicate either a lower
prioritization or greater technical difficulty, while the proliferation
of fairness and robustness tools may be driven by stronger regulatory
and public pressure, especially around concerns of algorithmic bias.

Educational tools aimed at building awareness, informing, preparing, or
upskilling stakeholders involved in or affected by AI systems include
training programs, online courses, best-practice guides, and other
resources designed to support trustworthy AI. However, these tools are
the least common overall, according to the OECD. Based on Figure~\ref{fig:tools-by-type}, the
limited number of educational tools tends to focus on broad principles
such as transparency, privacy, fairness, and human agency and control,
while areas like explainability, data governance, digital security, and
environmental sustainability are notably underserved. This points to a
significant capacity-building gap in AI ethics. While many organizations
appear to prioritize technical fixes or process-based approaches, there
has been far less investment in creating training modules or curricula
that educate practitioners and stakeholders about how to interpret model
outputs or secure AI systems. As a result, ethical AI training remains
largely non-institutionalized, and many developers, data scientists, and
other stakeholders may not be systematically educated on key ethical
issues. This aligns with findings in the literature. For instance, Munn
(2023) argues that the lack of ethics education contributes to AI
development in an ``ethically empty milieu'', where technologists are
unaware of, or unprepared for, the ethical standards expected of them.
The scarcity of educational tools underscores this problem and suggests
a need for more comprehensive ethics education and upskilling efforts in
the domain of AI.

Procedural tools, which offer operational or implementation guidance in
areas such as governance, risk management, and product development,
include resources like checklists, assessment frameworks, template
policies, and documentation standards that organizations can adopt to
embed ethical considerations into their workflows. These tools sit
between purely technical solutions and educational resources, providing
structured methods to integrate ethics into AI development and
organizational practices. The analysis shows that procedural tools most
frequently address transparency, fairness, and privacy, which aligns
with the suitability of documentation standards for transparency and
traceability, bias assessment checklists for fairness, and consent
checklists or data management policies for privacy. However, similar to
trends observed in other categories, procedural tools focused on
environmental sustainability, explainability, data governance, and
digital security are notably lacking. This suggests that standardized
processes in these domains, such as widely used checklists for ensuring
model explainability, templates for adversarial threat assessments, or
established protocols for evaluating the environmental footprint of AI
systems, have not yet been widely developed or adopted, at least as
reflected in the OECD catalogue. The absence of procedural tools for
these principles implies that such issues may be addressed through
ad-hoc methods or left solely to technical teams rather than being
supported by institutionalized processes. Since procedural tools play a
critical role in embedding ethical principles into everyday operations,
these gaps may hinder consistent and systematic implementation of
explainability, environmental sustainability, and security across AI
projects.

In summary, technical and procedural approaches currently dominate,
especially for well-known concerns like fairness, and transparency,
whereas educational approaches are underutilized as a strategy for
promoting ethical AI. This overreliance on tech and process, to the
neglect of training and culture, is a potential weakness. Experts argue
that cultivating an ethical mindset among AI practitioners is as
important as, or even more important than, any checklist or toolkit
(Mittelstadt 2019). Our findings support the idea that many AI ethics
initiatives provide little practical guidance, offering few concrete
recommendations or training materials to help put principles into
practice. As a result, designers and developers are often uncertain
about how to apply these principles in their work. Expanding educational
resources could help bridge the knowledge gap and empower a broader base
of stakeholders to engage with AI ethics, rather than assuming
technology alone can solve ethical issues. In this way, the human being
is affirmed as an active subject concerning technology, rather than
reduced to a passive object, thereby preserving agency and
responsibility in ethical engagement.

\subsection{Lifecycle Stage Focus: Design vs Post-Deployment}

\begin{figure}[tbp]
  \centering
  \includegraphics[width=\linewidth]{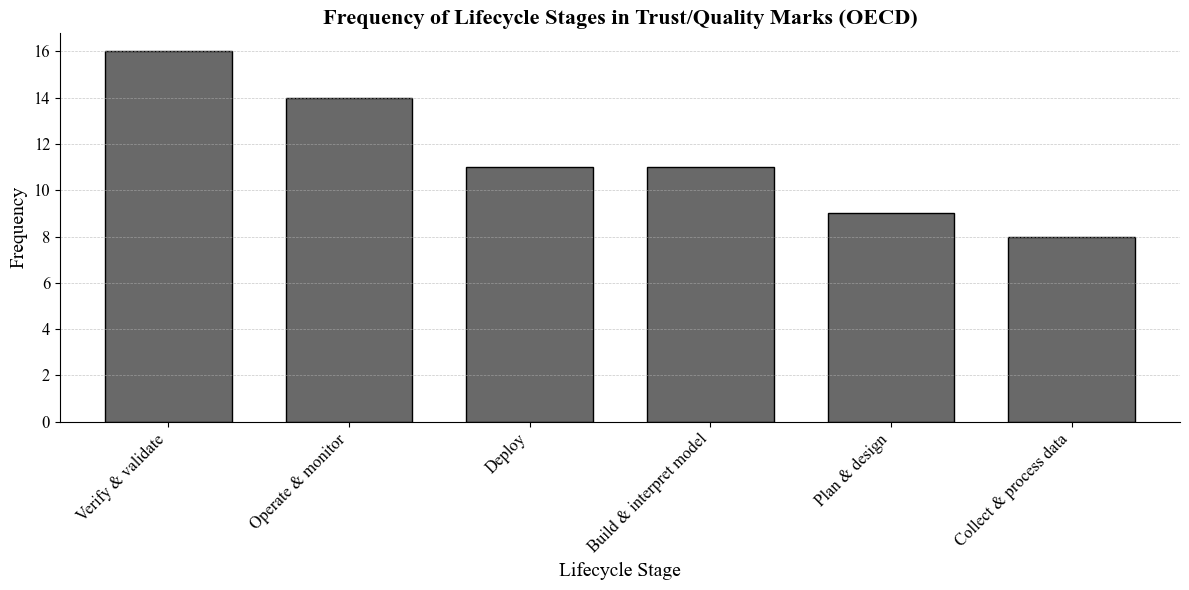}
  \caption{Number of trust/quality marks in the OECD catalogue by AI-system lifecycle phase.}
  \label{fig:marks-lifecycle}
\end{figure}

An important dimension of AI governance concerns the timing of ethical
interventions within the AI system lifecycle. The OECD's review of
trust/quality mark frameworks reveals a skewed focus toward the later
stages of AI development, with relatively little coverage of the early
phases. Specifically, the data show that these trust mark programs most
frequently address the ``Verify \& Validate'' and ``Operate \& Monitor''
stages of the AI lifecycle (Figure~\ref{fig:marks-lifecycle}). In other words, the bulk of
guidance and requirements in these frameworks take effect after an AI
model has been built, specifically during the phases of testing and
validation, as well as during deployment. This typically involves
activities like performance evaluation, bias auditing, safety
verification, and ongoing monitoring of the AI system's outputs and
impacts. This focus implies that trustworthiness is treated mainly as
something to be verified at the end of development or during operation,
similar to a final audit or compliance check.

Conversely, earlier lifecycle stages are comparatively neglected. Stages
such as ``Plan \& Design'' (the initial conception and design of the AI
system) and ``Collect \& Process Data'' (data gathering and preparation)
are mentioned much less frequently in the trust mark criteria. This
indicates a potential gap in early-stage guidance: few trustworthiness
schemes emphasize incorporating ethics by design at the project outset
or ensuring data ethics during dataset construction. The implication is
that many current certifications and marks assume the AI system is
already built, and then focus on evaluating its behavior and processes.
While post-hoc validation remains important, the ``Ethics by Design''
literature emphasizes that embedding ethical and societal considerations
early in system development leads to better outcomes by anticipatorily
integrating values into the design process and mitigating ethical risks
holistically (Winfield and Jirotka 2018; Umbrello and Van de Poel 2021;
Donia and Shaw 2021; European Commission 2021). Our finding that ``Plan
\& Design'' stage gets less attention aligns with critiques that much of
AI ethics has been reactive rather than proactive. Gianni et al. (2022)
note that although principles such as fairness are widely endorsed, it
is often unclear how they should be implemented in practice during
design, leaving developers puzzled by the lack of concrete direction.

This approach, where ethics are verified after development, can be
problematic. If ethical flaws, such as biased data or an unsafe model
architecture, are embedded early and only discovered later, it may be
too late or too costly to fix them. The strong focus on
validation/monitoring suggests that trust mark initiatives are currently
geared toward auditing and certifying AI systems at the point of use
rather than guiding their initial development. Indeed, AI auditing is
gaining prominence as a means to assess trustworthiness (Scharowski et
al. 2023). Audits and documentation reviews are being developed to
ensure compliance with the principles at deployment. However,
over-reliance on audits without ``Ethics by Design'' could result in a
check-box oriented mentality, where issues are addressed superficially
at the end. Our analysis underscores the need to extend ethical AI
practices to the design and data phases of the lifecycle. For example,
more trust marks could include criteria for requirements analysis with
ethical impact assessments, diverse and bias-aware data collection
strategies, or anthropocentric design processes. The limited coverage of
early stages in current marks indicates the community is still maturing
in this regard. Closing this chasm will likely involve developing new
standards and best practices specifically targeting the design stage,
which can then be incorporated into certification programs. In summary,
current trust marks focus on verification and pay relatively little
attention to guiding and supporting initial design phases. This finding
highlights an opportunity to push ethical considerations further
upstream in the AI lifecycle, complementing the existing emphasis on
end-stage accountability with earlier-stage ``Ethics by Design''
measures. Doing so would address the ``gulf between high-minded ideals
and technological development on the ground'' that has been observed in
AI ethics initiatives (Munn 2023).

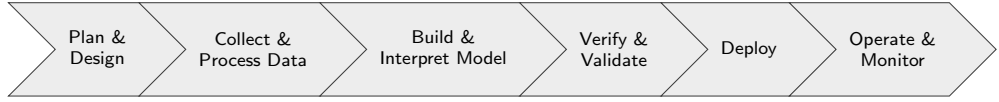
\begin{figure}[tbp]
  \centering
  \resizebox{\linewidth}{!}{%
  \begin{tikzpicture}[
    node distance=-0.8pt,
    every node/.style={font=\footnotesize\sffamily, align=center, minimum height=1.25cm},
    phase/.style={signal, signal from=west, signal to=east, draw=black!75,
      fill=black!7, minimum width=2.25cm, inner xsep=5pt}
  ]
    \node[phase] (p1) {Plan \&\\Design};
    \node[phase, right=of p1] (p2) {Collect \&\\Process Data};
    \node[phase, right=of p2] (p3) {Build \&\\Interpret Model};
    \node[phase, right=of p3] (p4) {Verify \&\\Validate};
    \node[phase, right=of p4] (p5) {Deploy};
    \node[phase, right=of p5] (p6) {Operate \&\\Monitor};
  \end{tikzpicture}%
  }
  \caption{Lifecycle phases of an AI system.}
  \label{fig:lifecycle}
\end{figure}

\subsection{Target Stakeholders: Technical Roles vs Policy Roles}

\begin{figure}[tbp]
  \centering
  \includegraphics[width=\linewidth]{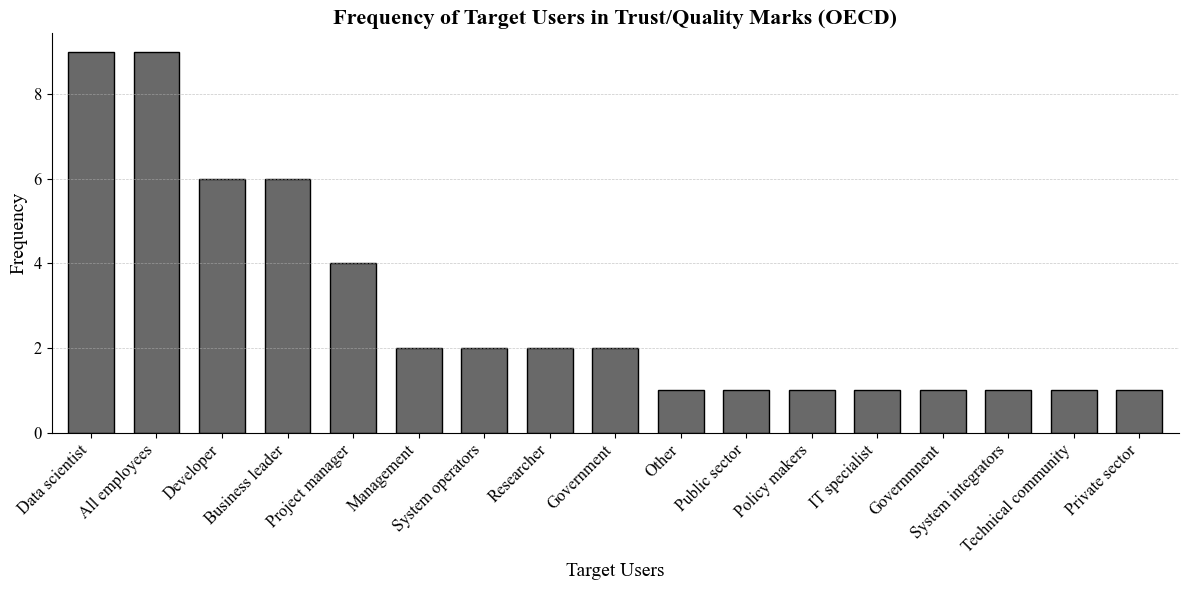}
  \caption{Number of target-user groups in the OECD catalogue for trust/quality marks.}
  \label{fig:target-users}
\end{figure}

Another dimension of the trust or quality mark frameworks analyzed is
the intended audience or target users of these guidelines and
certifications. The OECD catalogue indicates that the primary targets
are technical and industry roles directly involved in AI development and
deployment, rather than policymakers or general public stakeholders.
Referring to Figure~\ref{fig:target-users}, the most frequently mentioned user groups for
these trust mark frameworks are ``Data scientists'', ``Developers'', and
``Business leaders''. This suggests that the initiatives are mostly
geared towards those building or managing systems within organizations.
In many cases, the marks also commonly target ``All employees'' of an
organization, implying an internal focus on corporate or institutional
practice, ensuring everyone in a company follows certain AI quality
processes. This is reasonable, as many trustworthy AI certifications are
pursued by companies to signal that their AI products or processes meet
a standard, thereby involving internal compliance.

However, notably under-targeted are external and public sector
stakeholders: policymakers, regulators, the broader public sector, and
the research/technical community at large are much less frequently cited
as intended users. For example, few frameworks explicitly address the
needs of policymakers who might use such guidelines to craft
regulations, or public sector organizations deploying AI in government
services, or even the general public who might be affected by AI
decisions. This skew toward industry technical roles indicates that
current trust mark efforts are largely an industry-led self-governance
mechanism, rather than a tool for government oversight or public
empowerment. It also suggests a relatively narrow focus, as they aim to
instruct developers and companies on how to build trustworthy AI, but do
not necessarily help government bodies or civil society to evaluate or
demand trustworthiness in AI from the outside.

Many have noted that a large portion of AI principles has originated in
the private sector or through partnerships, with less representation
from civil society and some regions of the world (Jobin et al. 2019).
Our findings likewise indicate that AI ethics efforts are often
concentrated within technology organizations themselves. Gianni et al.
(2022) argue that official national strategies have tended to lean on
these industry-driven ethical guidelines, without sufficiently engaging
broader societal actors. In effect, there is a lack of multi-stakeholder
engagement, as current frameworks do not strongly involve policymakers
or the public in defining or applying trustworthiness criteria. The risk
is that important perspectives (e.g., human rights advocates, consumer
protection agencies, or marginalized user groups) might not be
adequately represented in how trustworthiness of AI systems or models is
operationalized. Indeed, Gianni et al. (2022) advocate for more
democratic and participatory approaches to AI governance, noting that
present guidelines often exclude contextual actors and thereby lack
legitimacy and efficacy. Furthermore, targeting only technical personnel
might limit the uptake of ethical practices. AI ethics is not just a
technical problem, but organizational leadership and policy environment
matter greatly (Morley et al. 2021). If a trust mark targets developers
rather than corporate boards or regulators, it may generate little
external pressure or institutional support to ensure the prescribed
practices are actually implemented. Some initiatives, like the EU AI
Act, aim to bring policymakers into the loop by mandating certain
compliance for high-risk or prohibited AI systems.

\begin{figure}[tbp]
  \centering
  \includegraphics[width=\linewidth]{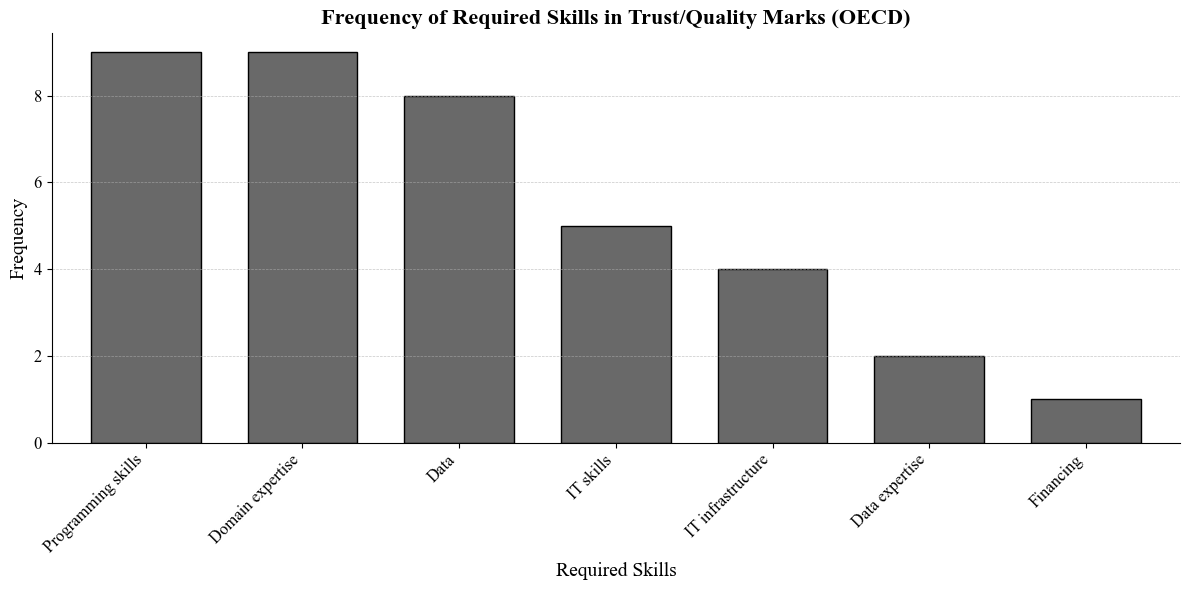}
  \caption{Number of required skills in the OECD catalogue for trust/quality marks.}
  \label{fig:required-skills}
\end{figure}

Overall, technical implementers are the primary audience of current
trustworthy AI marks, whereas policy-makers and public sector actors are
secondary or neglected audiences. This is also reflected in the skill
requirements (Figure~\ref{fig:required-skills}) specified in the OECD catalogue of trust/quality
mark frameworks, which emphasize competencies such as ``Programming
skills'', ``Data management'', and ``IT skills'', among others. This
indicates a need for broader outreach and inclusion. Future frameworks
might benefit from versions tailored to government and oversight use, or
collaborations that include public sector needs. Additionally,
interdisciplinary training could ensure that policymakers understand
technical criteria and developers understand regulatory expectations.
The present chasm between the technical and policy communities in AI
ethics, as evidenced by our findings, has been identified as a challenge
for governing AI responsibly and trustworthily (Morley et al. 2021;
Papademas et al. 2025). Bridging this gap will likely involve creating
channels for communication and co-development of standards between
engineers, social scientists, and policy officials. Synoptically, a
socio-technical approach is essential if the issue is to be addressed at
its root.

\begin{figure}[tbp]
  \centering
  \includegraphics[width=\linewidth]{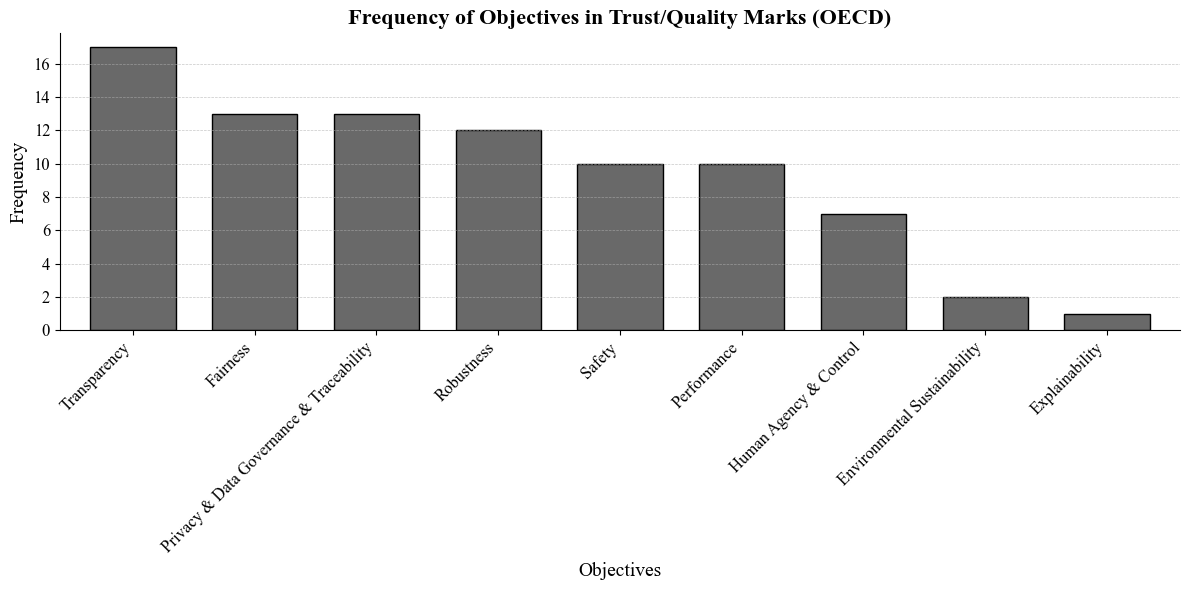}
  \caption{Number of trust/quality marks in the OECD catalogue per trustworthy-AI objective.}
  \label{fig:marks-objective}
\end{figure}

\begin{figure}[tbp]
  \centering
  \includegraphics[width=\linewidth]{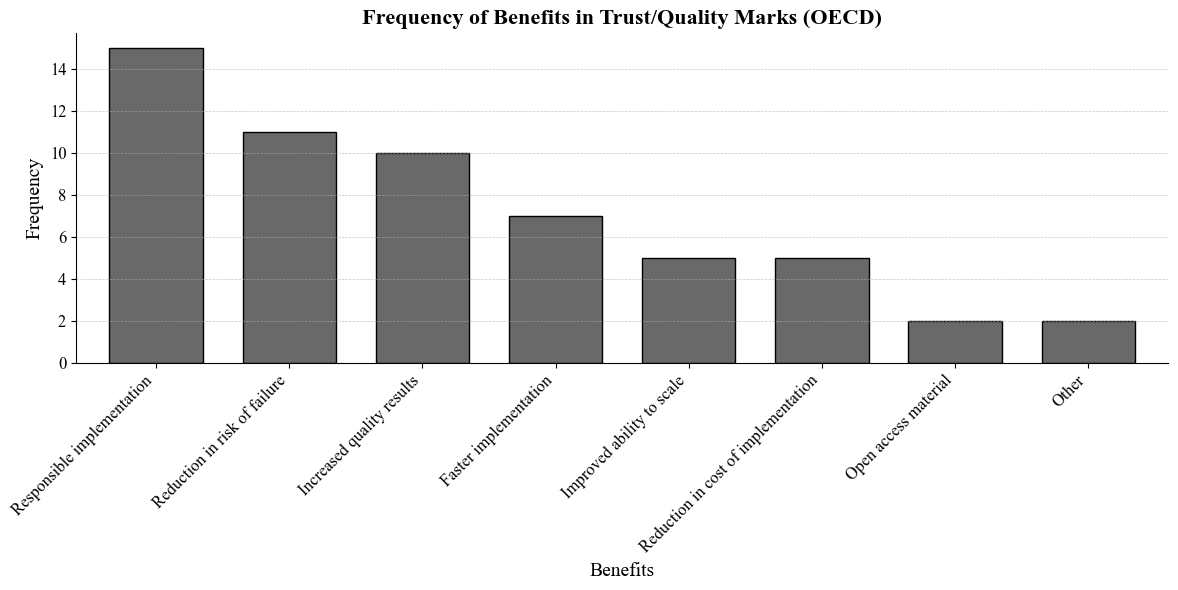}
  \caption{Number of trust/quality marks in the OECD catalogue per stated benefit.}
  \label{fig:marks-benefit}
\end{figure}

\section{Discussion}\label{sec:discussion}

The findings from the OECD's catalogue analysis align with key themes in
the scholarly discourse on AI ethics, while simultaneously offering
fresh insights into the current state of practice. In this section, we
interpret the results and consider their broader implications within the
context of existing research. Our observation that transparency,
fairness, and robustness/safety dominate both tool development and trust
framework criteria aligns with many SLRs, which report a growing
convergence around a few core ethical principles (Zeng et al. 2018;
Fjeld et al. 2020; Corrêa et al. 2023; Laine et al. 2024). This
convergence is encouraging insofar as it suggests a shared understanding
of fundamental goals for AI trustworthiness. This convergence likely
reflects the impact of prominent policy frameworks, such as the EU's
trustworthy AI requirements, which emphasize principles including
transparency, fairness/non-discrimination, and technical robustness. The
fact that these principles are now translating into concrete tools and
certification criteria is a positive sign of progress, indicating that
the abstract values are being operationalized to some extent. For
example, the plethora of fairness toolkits and bias audits shows the
principle of justice is being tackled in practice, and the emphasis on
transparency in frameworks and trust marks shows commitments to
trustworthiness are being formalized. This partially bridges the
principle-practice gap, answering calls, such as those in Mittelstadt
(2019), to move beyond merely listing principles.

However, underrepresented areas, such as explainability and
sustainability, highlight where convergence has not yet led to pragmatic
action. Sustainability, in particular, emerges as an area requiring
greater attention, as both our analysis and previous analyses (Corrêa et
al. 2023) consistently indicate that it is frequently overlooked. This
is concerning because the environmental footprint of AI, stemming from
energy-intensive training processes, resource consumption, and carbon
emissions, is substantial and directly intersects with questions of
ethical and responsible use. The lack of tools and criteria for
sustainability and environmental well-being suggests that ethical AI
efforts may be overly narrow, focusing solely on algorithmic fairness
and individual rights, without accounting for broader societal impacts
such as climate change. Similarly, explainability being neglected in
practice despite being frequently cited in principle lists, sometimes
under terms like ``explicability'' (Floridi and Cowls 2019), implies a
gap between rhetorical support and technical feasibility or commitment.
One interpretation is that organizations find transparency (making
information available) more feasible than true explainability (making AI
decisions understandable), and therefore tend to prioritize the former.
This gap aligns with the hypothesis that many AI explanations are
superficial and that genuine interpretability remains challenging. Our
findings underscore the need for research and investment in more
challenging areas, such as the development of robust Explainable AI
(XAI) (Freiesleben and König 2023) methods and their integration into
standards, as well as the creation of methodologies to quantify and
mitigate environmental impacts. Without addressing these thematics, the
current notion of trustworthy AI may achieve only a verisimilitude of
trustworthiness, remaining incomplete at its core.

\subsection{Technical and Procedural Dominance vs. Educational Needs}

The overwhelming reliance on technical/process solutions and the
rareness of educational initiatives is another key point. This mirrors a
classic pattern in technology governance where structural and
technological fixes are favored over human-centric solutions like
training and culture change. From a socio-scientific perspective, this
may limit the effectiveness of AI ethics programs. Research in
organizational behavior suggests that ethical outcomes depend greatly on
the ethical culture and competencies within an organization (Treviño et
al. 2006). If developers and managers lack ethical literacy, they might
implement checklists in a perfunctory way or find workarounds to pass
audits without genuine change, what some call ethicswashing or ethics
theater (Bauer and Lizotte 2021; Cath and Keys 2021; Cath and Jansen
2021). Munn (2023) argues that an ethically empty development
environment can persist despite principles on paper. Our finding that
training tools are scarce is a concrete datapoint supporting that
concern. It implies that many companies might not be investing in
upskilling their staff in AI ethics, instead relying on external tools
or consultants to assess ethics ex post. To improve this, integrating
ethics into education curricula for AI practitioners and providing
accessible organizational training modules is vital. Encouragingly, some
universities and companies are beginning to offer AI ethics courses and
explore the potential benefits and risks of AI (Dabis and Csáki 2024),
but these efforts need to be scaled up. It is recommended that future
trust frameworks include mechanisms to mandate or incentivize ethics
training. For instance, a certification could require that a certain
percentage of the AI team has undergone accredited ethics training, or
that the company runs annual workshops on trustworthy AI. Cultivating
critical thinking in individuals may provide a foundation for developing
trustworthy and ethical approaches to AI that are centered on human
values and interests. Such requirements would push organizations to
develop the human capacity for ethical thinking, not just the technical
capacity for ethical checking.

The aforementioned distinction is philosophically significant. A
governance ecosystem centered predominantly on tools and procedures
risks externalizing moral responsibility into artefacts of compliance,
thereby encouraging the view that ethical adequacy can be achieved
through the correct use of instruments alone. However, trustworthiness
is not an inherent feature of a technical system by itself. Instead, it
arises from socio-technical relationships involving designers,
institutions, affected communities, and the normative assumptions
embedded in practice. For this reason, educational deficits should be an
indicator that the cultivation of judgment, reflexivity, and practical
wisdom remains structurally undervalued within current trustworthy AI
initiatives and should not be interpreted as simply a lack of training
materials.

\subsection{Lifecycle Coverage - Embedding Ethics Early}

The bias towards later lifecycle phases in trust marks is indicative of
an audit culture, as opposed to a design culture. This has parallels in
other domains, e.g., software security, historically focused on
penetration testing (after development) until the rise of ``Secure by
Design'' philosophies. The literature on ethical design argues that many
concerns, such as algorithmic bias, are most effectively addressed
upstream through practices like inclusive design teams, conscientious
problem formulation (Passi and Barocas 2019), proactive data curation
(Andrews et al. 2023), and participatory design involving marginalized
groups (Newman-Griffis et al. 2022), rather than relying solely on
downstream audits. Our results suggest that the message has not yet
fully penetrated the certification landscape. The relative neglect of
``Plan \& Design'' guidance indicates a ripe area for development. We
foresee and recommend a shift: next-generation AI governance frameworks
should incorporate design-phase requirements. This could include
processes like algorithmic impact assessments conducted before model
development, participatory design with affected stakeholders, and
requirements that teams document how ethical considerations influenced
their design decisions (not just how the final model was evaluated). The
trick will be to create concrete standards for design practices that can
be verified. Our findings suggest a need for interdisciplinary work to
codify best practices for ethical AI design so that they can be adopted
by industrial frameworks.

The temporal imbalance identified here also points to a deeper issue
concerning when responsibility is recognized within technological
systems. If ethics enters primarily at the stages of validation and
monitoring, responsibility is framed retrospectively, as if the moral
quality of AI were something to be inspected after technical decisions
have already settled into infrastructures, datasets, and model
architectures. A more adequate socio-technical understanding would treat
normativity as constitutive rather than corrective, as values are not
appended to systems at the end of development, but are already inscribed
in problem formulation, design choices, data selection, and assumptions
about users and harms. Thus, the limited attention to early stages
reveals a particular moral chronology in which ethical reflection is
displaced downstream, where it is often weaker, more expensive, and less
transformative.

\subsection{Stakeholder Engagement and Audience Gaps}

The finding that policymakers and external stakeholders are not primary
targets raises concerns about who drives AI ethics. If it remains an
internal matter within technology companies, there is a significant risk
of compromised oversight, as firms would effectively be assessing their
own adherence to ethical standards. The limited involvement of
regulators or independent oversight in these frameworks might allow
self-satisfaction or superficial compliance. This is precisely why many
researchers argue for co-regulation or independent audits (Falco et al.
2021; Raji et al. 2022; OECD 2024; Li and Goel 2025). Our data suggests
that trust marks are being used as self-policing tools by companies,
rather than as independent mechanisms. Integrating policymakers could
take the form of public-private partnerships developing standards,
similar to how the financial industry has both internal controls and
external audit/regulation. Since our analysis indicates that existing
certification schemes are not primarily directed at governments,
extending their scope to cover public-sector adoption could
significantly strengthen the trustworthiness of the AI entities.
Additionally, the limited focus on the public sector is notable because
government AI use has a significant societal impact and arguably
warrants as much, if not greater, oversight than private sector use.
This gap might partly stem from the fact that many frameworks were
developed in corporate contexts with less input from public agencies.
Researchers have examined how national AI strategies incorporate ethical
considerations (Ulnicane et al. 2021), highlighting considerable
variation across countries. Our findings further suggest that, at the
global level, the connection between voluntary standards and concrete
policy action remains tenuous. Strengthening this linkage, whether
through the formal adoption of standards or explicit regulatory
references, will be crucial to ensure that ethics frameworks function as
more than voluntary guidelines. The stakeholder asymmetry identified can
also be read as an epistemic authority asymmetry. When trustworthiness
is primarily articulated for developers, data scientists, and business
actors, the power to define what counts as a trustworthy system remains
concentrated among those already positioned closest to production and
deployment. This asymmetry risks narrowing the normative horizon of
governance, because communities affected by AI systems often experience
harms, dependencies, and exclusions that are not fully visible from
within technical or managerial standpoints. A genuinely trustworthy
socio-technical order requires a redistribution of interpretive
authority over how risks, benefits, and acceptable trade-offs are
understood.

In essence, our discussion highlights that while the current ecosystem
of AI ethics tools and marks is a significant step beyond mere
principles on paper, it still reflects early-stage governance, with
various imbalances to address. The trends we found echo known issues,
such as the lack of enforcement and the need for education, that
scholars have been raising. The contribution of this analysis is to
pinpoint exactly where those issues manifest in the practical landscape.
It provides evidence that, for example, explainability is not merely a
theoretical concern but is, in fact, lacking in existing tools and
standards. By matching these findings with the literature, we see
consistency: many chasms in implementation correspond to critiques of
the literature on vagueness, lack of teeth, and the narrow focus of AI
ethics efforts. On a positive note, the areas of strength (transparency,
fairness, etc.) show that concerted effort can yield tools and
consensus. If similar effort is now applied to the weaker areas, the
state of trustworthy AI governance could become more comprehensive.
Taken together, these findings suggest that the current landscape of
trustworthy AI implementation is structured by a tension between ethical
breadth at the level of principle and ethical selectivity at the level
of practice. The OECD ecosystem shows that implementation does not
simply execute pre-given moral commitments, but filters them through
organizational feasibility, technical legibility, and institutional
incentives. In this respect, the contribution of the present study is
not limited to confirming that familiar concerns persist.
Antithetically, it empirically demonstrates where and how the
contemporary infrastructure of trustworthy AI narrows, translates, and
stabilizes particular ethical priorities while marginalizing others. The
importance of this observation is socio-technical as much as
philosophical, as it indicates that the future of AI governance will
depend not only on formulating better principles, but also on
interrogating the conditions under which certain values become
implementable, auditable, and governable in the first place.

\subsection{Recommendations}

Building on both our findings and broader research, we propose
recommendations to advance trustworthy AI governance. Ethical objectives
for tools and trust/quality marks should be broadened to encompass
environmental sustainability, explainability, and other currently
underrepresented principles, guided by domain-specific research, such as
work on XAI, to establish robust benchmarks. Ethics should be integrated
earlier in the development process, for instance, through design-stage
checklists or mandatory ethical risk assessments before deployment as
part of certification. Inclusivity must be enhanced by engaging
policymakers, end-user representatives, and interdisciplinary experts in
the development and application of trust frameworks, thereby ensuring
diverse perspectives are embedded. Organizations should be required to
invest in ethical competence through education and training, supported
by the sharing of best practices, which contributes to long-term ethical
capacity and enhances critical thinking in issues related to AI
trustworthiness. Enforcement mechanisms also need to be strengthened by
linking voluntary trust marks to regulatory or contractual requirements,
establishing independent audit capabilities to verify compliance, and
developing algorithmic or automated methods of assessment that can
efficiently elicit and evaluate the trustworthiness of AI systems.
Furthermore, governance frameworks should be regularly updated through
structured processes such as annual reviews or modular revisions,
enabling them to adapt to technological advances and new societal
insights. These recommendations align with scholarly calls to move from
abstract principles to pragmatic operationalization (Schiff et al.
2020b; Morley et al. 2021; Schiff et al. 2021; Morley et al. 2023), and
our study provides concrete evidence to support this transition,
emphasizing that the evolution of trustworthy AI governance is a
continuous process that must keep pace with innovation and societal
expectations.

\section{Conclusion}\label{sec:conclusion}

In this paper, we present an analytical examination of the current
landscape of trustworthy AI tools and trust/quality mark frameworks,
grounded in empirical findings from the OECD catalogue and supported by
relevant literature. Our analysis reveals a nuanced picture: the AI
community has made significant strides in developing tools and standards
for trustworthy AI, particularly around widely endorsed principles like
transparency, fairness, and robustness. There is clear evidence of the
translation of abstract principles into practice. For example, numerous
technical toolkits have been developed to measure bias and enhance model
transparency, and organizations are increasingly adopting trust mark
schemes to demonstrate their commitment to trustworthy AI. These
developments exhibit an important evolution from the early days of AI
ethics, when high-level principles were plentiful but concrete actions
were scarce. They reflect growing maturity and consensus in certain
areas of AI governance. At the same time, our study highlights critical
gaps and imbalances that need to be addressed to achieve a truly
comprehensive and effective AI governance regime. We found that aspects
such as explainability, digital security, and environmental
sustainability are under-prioritized in both tool development and trust
marks. This indicates that current efforts may be too narrowly focused,
leaving significant ethical dimensions at the margins. Bridging this
chasm will likely require deliberate efforts by researchers,
practitioners, and standard-setters to expand the suite of tools,
guidelines, and practices addressing these issues. For example, more
investment in explainable AI techniques and their integration into
practice is needed so that future certifications can demand a level of
interpretability from AI systems. Similarly, incorporating environmental
sustainability metrics into AI project evaluations could ensure that
environmental concerns become a standard part of AI ethics assessments.

We also identified a skew in the types of interventions being deployed.
Technical and procedural approaches dominate, while educational
initiatives are few. This imbalance highlights a potential weakness in
current strategies, as the human and organizational dimensions of AI
ethics are not being adequately addressed or cultivated. Building an
ethical AI culture is as important as any technical fix. Our findings
support that view and suggest that organizations and consortia should
place greater emphasis on ethics training and capacity-building.
Incorporating mandatory training elements into trustworthy AI programs
could help ensure that ethical principles are internalized by AI
developers and decision-makers, rather than treated as external
checkpoints. Over time, a workforce educated in AI ethics can make more
informed, conscious choices at each step of AI development, potentially
reducing the burden on end-of-life audits. Another insight from our
analysis is a mismatch in lifecycle focus: trust frameworks focus on
post-development verification and relatively neglect the design phase.
This indicates a reactive posture prevalent in current implementations.
To essentially achieve ``Ethics by Design'', the field must shift some
focus to earlier stages. This could involve developing new standards or
marks to certify ethical design processes or ethical data-sourcing
practices. For example, a certification might require evidence that an
AI team conducted an algorithmic impact assessment (Reisman et al. 2018;
Metcalf et al. 2021) at project inception, engaged with affected
stakeholders, or followed guidelines for inclusive dataset construction.
Our study provides impetus for such measures, aligning with calls in the
literature to embed ethics from the start. By tackling issues early on,
we can prevent or reduce challenges that might become deeply rooted and
costly to fix later, such as structural biases in training data or
opaque model logic that undermines accountability. Taking a proleptic
approach at the design stage also helps align system objectives with
societal values, fostering more trustworthy and sustainable AI outcomes.
The stakeholder targeting gap we uncovered, namely the focus on
technical roles contrasted with the minimal engagement of policymakers
and the public, is also significant. It suggests that current ethical AI
efforts are somewhat siloed within the tech industry. Remedying this
will require multi-stakeholder engagement, involving policymakers,
regulators, end-user communities, and domain experts in the creation and
governance of trust frameworks. It may also require adapting the
communication of these frameworks to make them accessible and useful to
non-technical stakeholders. The benefits of such an approach are
twofold: it promotes wider stakeholder engagement, thereby strengthening
the legitimacy and adoption of the frameworks, and it facilitates the
integration of diverse values and concerns that may otherwise be
overlooked by exclusively technical groups. In practice, this could mean
joint initiatives between industry consortia and government agencies to
co-develop certification criteria, or public consultation processes for
setting AI ethics standards. Our findings highlight the critical
importance of this trajectory, as without substantive policy-level
integration, ethics marks risk remaining voluntary and superficial,
lacking the enforceability, accountability, and institutional support
required to ensure meaningful and lasting influence on AI governance
practices.

In conclusion, the state of ethics and trust standards for AI, as
illuminated by the OECD findings, is one of significant progress coupled
with significant room for improvement. There is clear movement beyond
mere rhetoric: tools are being built, and standards are being applied,
an indicator of the dedication of many in the AI and policy communities
over the past few years. Yet, to ensure these efforts truly foster
trustworthy AI that is safe, fair, and beneficial for all, several
chasms must be addressed. More specifically, we elicit that the
implementation ecosystem of trustworthy AI does not simply translate
ethical principles into practice, but selectively stabilizes those
values that can be more readily formalized, measured, and governed
within existing socio-technical arrangements. Furthermore, our
perspective on the OECD's data underscores several critical directions
for the future: expanding the ethical framework to encompass
environmental sustainability as well as the explainability and
interpretability of AI systems; embedding ethical considerations
throughout the entire AI lifecycle, thereby demonstrating the value of
an ``Ethics by Design'' approach; and the cultivation of polyphonic
participation from technical experts to social actors, and from
individual agents to collective bodies.

The insights from this study can inform both practitioners and
researchers. For practitioners and policymakers designing the next
generation of AI governance mechanisms, our results highlight priority
areas, such as explainability and environmental sustainability, that
merit greater attention, and caution against over-reliance on
self-reporting and post hoc fixes. For researchers, the findings point
to underexplored empirical questions (e.g., how to measure and improve
the effectiveness of educational interventions in AI ethics, or how
trust marks influence AI developer behavior in practice) and invite
continued evaluation of whether implemented tools are achieving the
intended ethical outcomes. As AI systems become ever more entwined with
society, the importance of such rigorous, evidence-backed approaches to
ethics and governance will only grow. We hope that this work contributes
to an ongoing dialogue between empirical analysis and normative
scholarship, ultimately supporting the development of AI that not only
functions effectively in a technical sense but also earns the trust of
those it affects, thereby striving toward a synthesis of innovation and
moral responsibility. In this sense, the central challenge is to
transform the institutional, epistemic, and design conditions under
which AI ethics principles become materially operative, without
following the tactic of producing a plethora of theoretical principles
that are not accompanied by techniques for implementing them in the
pragmatic world.

\section*{Author contributions}

Michael Papademas conducted the analysis, developed the methodology, and
wrote the manuscript. Xenia Ziouvelou contributed to the design of the
methodology, reviewed the work, and supervised the study. Kostas Karpouzis
reviewed the work and supervised the study. Vangelis Karkaletsis supervised
the study.

\section*{Funding}

The author(s) declare that financial support was received for the research
and/or publication of this article. This work has been funded by the Digital
Europe Programme (DIGITAL) under Grant Agreement No.~101146490 --
DIGITAL-2022-CLOUD-AI-\mbox{B-03}.

%
%
%

\section*{References}
\addcontentsline{toc}{section}{References}

\refentry{Andrews, J., Zhao, D., Thong, W., Modas, A., Papakyriakopoulos, O., \& Xiang, A. (2023). Ethical considerations for responsible data curation. Advances in Neural Information Processing Systems, 36, 55320--55360.}

\refentry{Bauer, G. R., \& Lizotte, D. J. (2021). Artificial intelligence, intersectionality, and the future of public health. American Journal of Public Health, 111(1), 98--100. \url{https://doi.org/10.2105/AJPH.2020.306006}}

\refentry{Bietti, E. (2020). From ethics washing to ethics bashing: A view on tech ethics from within moral philosophy. In Proceedings of the 2020 Conference on Fairness, Accountability, and Transparency (pp. 210--219). Association for Computing Machinery. \url{https://doi.org/10.1145/3351095.3372860}}

\refentry{Cath, C., \& Jansen, F. (2021). Dutch comfort: The limits of AI governance through municipal registers. arXiv Preprint arXiv:2109.02944. \url{https://arxiv.org/abs/2109.02944}}

\refentry{Cath, C., \& Keyes, O. (2021). Your thoughts for a penny? Capital, complicity and AI ethics. In T. Phan, J. Goldenfein, D. Kuch, \& M. Mann (Eds.), Economies of virtue (Vol. 28, pp. 24--38). \url{http://dx.doi.org/10.25969/mediarep/19267}}

\refentry{Celsi, L. R., \& Zomaya, A. Y. (2025). Perspectives on managing AI ethics in the digital age. Information, 16(4), 318. \url{https://doi.org/10.3390/info16040318}}

\refentry{Corrêa, N. K., Galvão, C., Santos, J. W., Del Pino, C., Pinto, E. P., Barbosa, C., Massmann, D., Mambrini, R., Galvão, L., Terem, E., \& de Oliveira, N. (2023). Worldwide AI ethics: A review of 200 guidelines and recommendations for AI governance. Patterns, 4(10), 100857. \url{https://doi.org/10.1016/j.patter.2023.100857}}

\refentry{Dabis, A., \& Csáki, C. (2024). AI and ethics: Investigating the first policy responses of higher education institutions to the challenge of generative AI. Humanities and Social Sciences Communications, 11(1), 1--13. \url{https://doi.org/10.1057/s41599-024-03526-z}}

\refentry{Donia, J., \& Shaw, J. A. (2021). Ethics and values in design: A structured review and theoretical critique. Science and Engineering Ethics, 27(5), 57. \url{https://doi.org/10.1007/s11948-021-00329-2}}

\refentry{European Commission. (2018). Ethics guidelines for trustworthy AI. \url{https://digital-strategy.ec.europa.eu/en/library/ethics-guidelines-trustworthy-ai}}

\refentry{European Commission. (2020). Directorate-General for Communications Networks, Content and Technology. The Assessment List for Trustworthy Artificial Intelligence (ALTAI) for self assessment. Publications Office. \url{https://data.europa.eu/doi/10.2759/002360}}

\refentry{European Commission. (2021). Ethics by design and ethics of use approaches for artificial intelligence. \url{https://ec.europa.eu/info/funding-tenders/opportunities/docs/2021-2027/horizon/guidance/ethics-by-design-and-ethics-of-use-approaches-for-artificial-intelligence_he_en.pdf}}

\refentry{Floridi, L., \& Cowls, J. (2019). A Unified Framework of Five Principles for AI in Society. Harvard Data Science Review, 1(1). \url{https://doi.org/10.1162/99608f92.8cd550d1}}

\refentry{Fjeld, J., Achten, N., Hilligoss, H., Nagy, A., \& Srikumar, M. (2020). Principled artificial intelligence: Mapping consensus in ethical and rights-based approaches to principles for AI. SSRN Electronic Journal. \url{https://doi.org/10.2139/ssrn.3518482}}

\refentry{Freiesleben, T., \& König, G. (2023). Dear XAI community, we need to talk! Fundamental misconceptions in current XAI research. In Proceedings of the World Conference on Explainable Artificial Intelligence (pp. 48--65). Springer.}

\refentry{Gianni, R., Lehtinen, S., \& Nieminen, M. (2022). Governance of responsible AI: From ethical guidelines to cooperative policies. Frontiers in Computer Science, 4, 873437. \url{https://doi.org/10.3389/fcomp.2022.873437}}

\refentry{Jobin, A., Ienca, M., \& Vayena, E. (2019). The global landscape of AI ethics guidelines. Nature Machine Intelligence, 1, 389--399. \url{https://doi.org/10.1038/s42256-019-0088-2}}

\refentry{Klein, L., \& D'Ignazio, C. (2024). Data feminism for AI. In Proceedings of the 2024 ACM Conference on Fairness, Accountability, and Transparency (pp. 100--112). \url{https://doi.org/10.1145/3630106.3658543}}

\refentry{Laine, J., Minkkinen, M., \& Mäntymäki, M. (2024). Ethics-based AI auditing: A systematic literature review on conceptualizations of ethical principles and knowledge contributions to stakeholders. Information \& Management, 61(5), 103969. \url{https://doi.org/10.1016/j.im.2024.103969}}

\refentry{Li, Y., \& Goel, S. (2025). Making it possible for the auditing of AI: A systematic review of AI audits and AI auditability. Information Systems Frontiers, 27, 1121--1151. \url{https://doi.org/10.1007/s10796-024-10508-8}}

\refentry{Metcalf, J., Moss, E., Watkins, E. A., Singh, R., \& Elish, M. C. (2021). Algorithmic impact assessments and accountability: The co-construction of impacts. In Proceedings of the 2021 ACM Conference on Fairness, Accountability, and Transparency (pp. 735--746). Association for Computing Machinery. \url{https://doi.org/10.1145/3442188.3445935}}

\refentry{Metzinger, T. (2019). Ethics washing made in Europe. Der Tagesspiegel. \url{https://www.tagesspiegel.de/politik/eu-guidelines-ethics-washing-made-in-europe/24195496.html}}

\refentry{Mittelstadt, B. (2019). Principles alone cannot guarantee ethical AI. Nature Machine Intelligence, 1, 501--507. \url{https://doi.org/10.1038/s42256-019-0114-4}}

\refentry{Morley, J., Elhalal, A., Garcia F., Kinsey, L., Mökander, J., \& Floridi, L. (2021). Ethics as a service: A pragmatic operationalisation of AI ethics. Minds \& Machines, 31, 239--256. \url{https://doi.org/10.1007/s11023-021-09563-w}}

\refentry{Morley, J., Kinsey, L., Elhalal, A., Garcia F., Marta, Z., \& Floridi, L. (2023). Operationalising AI ethics: Barriers, enablers and next steps. AI \& Society, 38, 411--423. \url{https://doi.org/10.1007/s00146-021-01308-8}}

\refentry{Munn, L. (2023). The uselessness of AI ethics. AI and Ethics, 3(3), 869--877. \url{https://doi.org/10.1007/s43681-022-00209-w}}

\refentry{Newman-Griffis, D., Rauchberg, J. S., Alharbi, R., Hickman, L., \& Hochheiser, H. (2022). Definition drives design: Disability models and mechanisms of bias in AI technologies. arXiv Preprint arXiv:2206.08287. \url{https://arxiv.org/abs/2206.08287}}

\refentry{OECD. (n.d.). OECD AI principles. \url{https://www.oecd.org/en/topics/ai-principles.html}}

\refentry{OECD. (2024). Enhancing co-operation between internal and external auditors: Towards a well-co-ordinated and strengthened public sector audit to ensure public accountability. Policy Paper. \url{https://doi.org/10.1787/0d4976ed-en}}

\refentry{Papademas, M., Ziouvelou, X., Troumpoukis, A., \& Karkaletsis, V. (2025). Bridging ethical principles and algorithmic methods: An alternative approach for assessing trustworthiness in AI systems. Frontiers in Computer Science, 7, 1658128. \url{https://doi.org/10.3389/fcomp.2025.1658128}}

\refentry{Passi, S., \& Barocas, S. (2019). Problem formulation and fairness. In Proceedings of the Conference on Fairness, Accountability, and Transparency (pp. 39--48). Association for Computing Machinery. \url{https://doi.org/10.1145/3287560.3287567}}

\refentry{Raji, I. D., Xu, P., Honigsberg, C., \& Ho, D. (2022). Outsider oversight: Designing a third-party audit ecosystem for AI governance. In Proceedings of the 2022 AAAI/ACM Conference on AI, Ethics, and Society (pp. 557--571). Association for Computing Machinery. \url{https://doi.org/10.1145/3514094.3534181}}

\refentry{Rees, C., \& Müller, B. (2023). All that glitters is not gold: Trustworthy and ethical AI principles. AI and Ethics, 3(4), 1241--1254. \url{https://doi.org/10.1007/s43681-022-00232-x}}

\refentry{Reisman, D., Schultz, J., Crawford, K., \& Whittaker, M. (2018). Algorithmic impact assessments: A practical framework for public agency. AI Now Institute Report. \url{https://ainowinstitute.org/publications/algorithmic-impact-assessments-report-2}}

\refentry{Rességuier, A., \& Rodrigues, R. (2020). AI ethics should not remain toothless! A call to bring back the teeth of ethics. Big Data \& Society, 7(2), 2053951720942541. \url{https://doi.org/10.1177/2053951720942541}}

\refentry{Rohde, F., Wagner, J., Meyer, A., Reinhard, P., Voss, M., Petschow, U., \& Mollen, A. (2023). Broadening the perspective for sustainable AI: Comprehensive sustainability criteria and indicators for AI systems. arXiv Preprint arXiv:2306.13686. \url{https://doi.org/10.48550/arXiv.2306.13686}}

\refentry{Ryan, M., \& Stahl, B. C. (2020). Artificial intelligence ethics guidelines for developers and users: Clarifying their content and normative implications. Journal of Information, Communication and Ethics in Society, 19(1), 61--86. \url{https://doi.org/10.1108/JICES-12-2019-0138}}

\refentry{Scharowski, N., Benk, M., Kühne, S. J., Wettstein, L., \& Brühlmann, F. (2023). Certification labels for trustworthy AI: Insights from an empirical mixed-method study. In Proceedings of the 2023 ACM Conference on Fairness, Accountability, and Transparency (pp. 248--260). Association for Computing Machinery. \url{https://doi.org/10.1145/3593013.3593994}}

\refentry{Schiff, D., Biddle, J., Borenstein, J., \& Laas, K. (2020a). What's next for AI ethics, policy, and governance? A global overview. In Proceedings of the AAAI/ACM Conference on AI, Ethics, and Society (pp. 153--158). Association for Computing Machinery. \url{https://doi.org/10.1145/3375627.3375804}}

\refentry{Schiff, D., Rakova, B., Ayesh, A., Fanti, A., \& Lennon, M. (2020b). Principles to practices for responsible AI: Closing the gap. arXiv Preprint arXiv:2006.04707. \url{https://doi.org/10.48550/arXiv.2006.04707}}

\refentry{Schiff, D., Rakova, B., Ayesh, A., Fanti, A., \& Lennon, M. (2021). Explaining the principles to practices gap in AI. IEEE Technology and Society Magazine, 40(2), 81-94. \url{https://doi.org/10.1109/MTS.2021.3056286}}

\refentry{Schultz, M. D., Conti, L. G., \& Seele, P. (2024). Digital ethicswashing: A systematic review and a process-perception-outcome framework. AI and Ethics, 5(2), 805--818. \url{https://doi.org/10.1007/s43681-024-00430-9}}

\refentry{Treviño, L. K., Weaver, G. R., \& Reynolds, S. J. (2006). Behavioral ethics in organizations: A review. Journal of Management, 32(6), 951--990. \url{https://doi.org/10.1177/0149206306294258}}

\refentry{Ulnicane, I., Knight, W., Leach, T., Stahl, B. C., \& Wanjiku, W. G. (2021). Framing governance for a contested emerging technology: Insights from AI policy. Policy and Society, 40(2), 158--177. \url{https://doi.org/10.1080/14494035.2020.1855800}}

\refentry{Umbrello, S., \& van de Poel, I. (2021). Mapping value sensitive design onto AI for social good principles. AI and Ethics, 1(3), 283--296. \url{https://doi.org/10.1007/s43681-021-00038-3}}

\refentry{Van den Bergh, J., \& Deschoolmeester, D. (2010). Ethical decision making in ICT: Discussing the impact of an ethical code of conduct. Communications of the IBIMA, 1. \url{https://doi.org/10.5171/2010.127497}}

\refentry{Winfield, A. F. T., \& Jirotka, M. (2018). Ethical governance is essential to building trust in robotics and artificial intelligence systems. Philosophical Transactions of the Royal Society A, 376(2133), 20180085. \url{https://doi.org/10.1098/rsta.2018.0085}}

\refentry{Zeng, Y., Lu, E., \& Huangfu, C. (2018). Linking artificial intelligence principles. arXiv Preprint arXiv:1812.04814. \url{https://doi.org/10.48550/arXiv.1812.04814}}

\end{document}